\documentclass[a4paper, conference]{Setup/ieeeconf}
\usepackage[english]{babel}
\usepackage{amsmath, amsfonts, bm}

\usepackage[
backend=biber,
style=ieee,
sorting=none,
doi=false,
isbn=false,
url=false,
eprint=false
]{biblatex}
\usepackage{graphicx}
\usepackage{float}
\usepackage{caption}
\usepackage{subcaption}
\usepackage{todonotes}
\usepackage{accents}
\usepackage[export]{adjustbox}

\DeclareSymbolFont{symbolsC}{U}{txsyc}{m}{n}
\DeclareMathSymbol{\circleleft}{\mathrel}{symbolsC}{146}
\DeclareMathSymbol{\circleright}{\mathrel}{symbolsC}{145}

\DeclareRobustCommand{\rightgroupderiv}[1]{\accentset{\scriptscriptstyle{\circleright}}{#1}}

\newcommand{\Adjoint}[1]{\textrm{Ad}_{#1}}
\newcommand{\Adjointinv}[1]{\textrm{Ad}_{#1}^{\textrm{-}1}}
\newcommand{\Adjointinvtpose}[1]{\textrm{Ad}_{#1}^{\textrm{-}T}}

\IEEEoverridecommandlockouts                              

\title{\LARGE \bf
Linear Kinematics for General Constant Curvature and Torsion Manipulators}

\author{Bill Fan$^{*1}$ Farhan Rozaidi$^{2}$, Capprin Bass$^{2}$, Gina Olson$^{3}$, Melinda Malley$^{1}$, Ross L Hatton$^{2}$
\thanks{*: Corresponding author}
\thanks{$^{1}$ Bill Fan and Melinda Malley are with Olin College of Engineering, Needham, MA, USA.
        {\tt\small \{wfan, mmalley\}@olin.edu}}%
\thanks{$^{2}$Farhan Rozaidi, Capprin Bass, and Ross L Hatton are with the Collaborative Robotics and Intelligent Systems (CoRIS) Institute, Oregon State University, Corvallis, OR, USA.
        {\tt\small \{nikahman, basscap, ross.hatton\}@oregonstate.org}}
\thanks{$^{3}$ Gina Olson is with the University of Massachusetts Amherst, Amherst, MA, USA.
        {\tt\small ginaolson@umass.edu}}%
}

\begin{document}

\maketitle
\thispagestyle{empty}
\pagestyle{empty}

\begin{abstract}
We present a novel general model that unifies the kinematics of constant curvature and constant twist continuum manipulators. Combining this kinematics with energy-based physics, we derive a linear mapping from actuator configuration to manipulator deformation that is analogous to traditional robot forward kinematics. Our model generalizes across manipulators with different sizes, types of bending, and types of actuators, without the need for parameter re-fitting. The combination of generality and linearity makes the model useful for control and planning algorithms. Finally, our model is shown to be accurate through experimental validation on manipulators with pneumatic artificial muscles.

\end{abstract}

\section{INTRODUCTION}

While the motion of traditional robots comes from their discrete joints, a continuum manipulator moves by deforming along its entire arc. These manipulators are often composed of rigid skeletons and soft actuators. When the actuators are activated, the interplay of forces within the manipulator creates stretching, bending, and sometimes twisting of the whole manipulator. These actuators commonly take the form of tendons or wires actuated by pulleys, pneumatic or hydraulic artificial muscles which contract when inflated, or shaped metal alloys which contract when electricity is applied.

%

\begin{figure}
    \centering
    \begin{subfigure}[t]{0.32\columnwidth}
        \includegraphics[width=\textwidth]{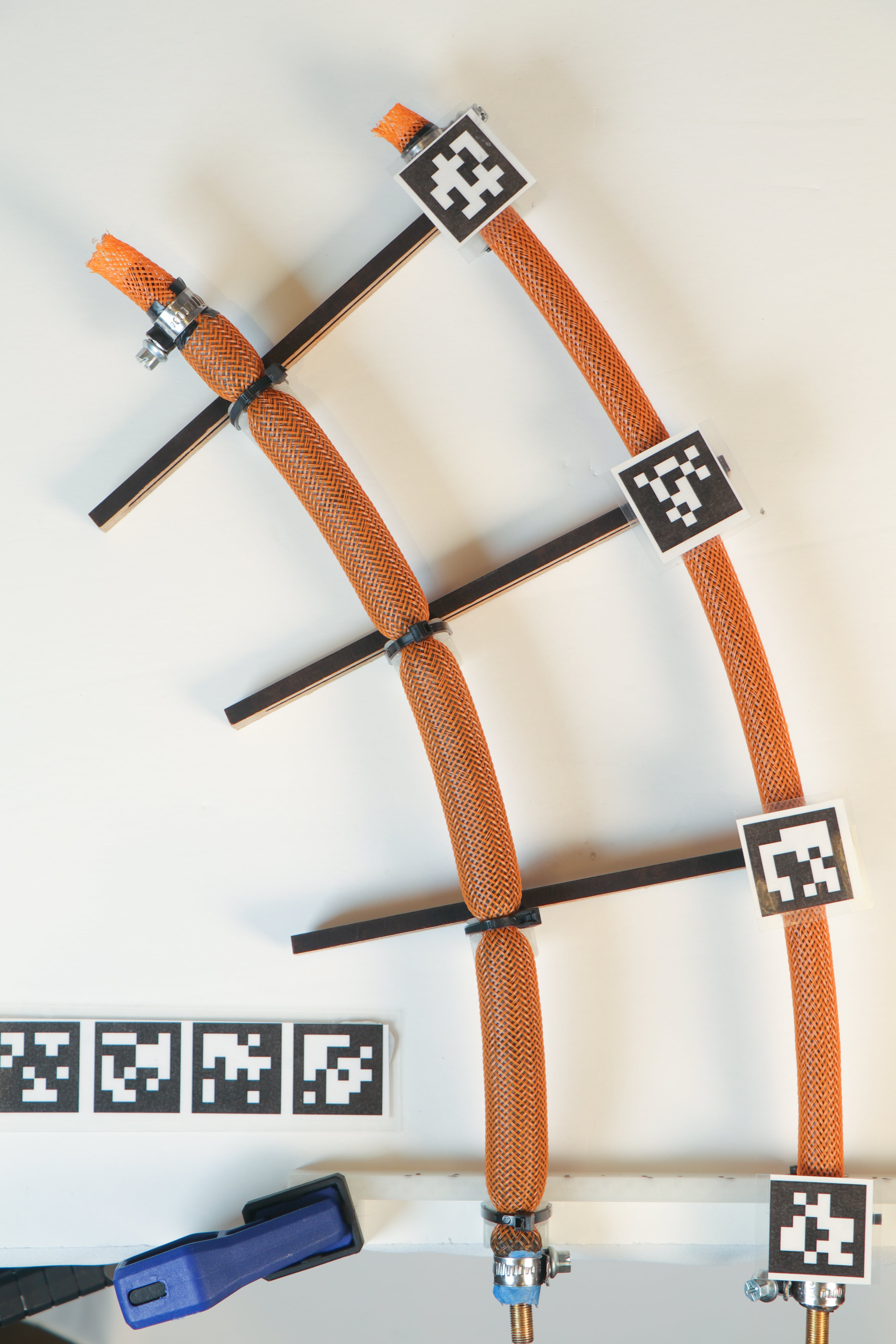}
        \centering
    \end{subfigure}
    \hfill
    \begin{subfigure}[t]{0.32\columnwidth}
        \includegraphics[width=\textwidth]{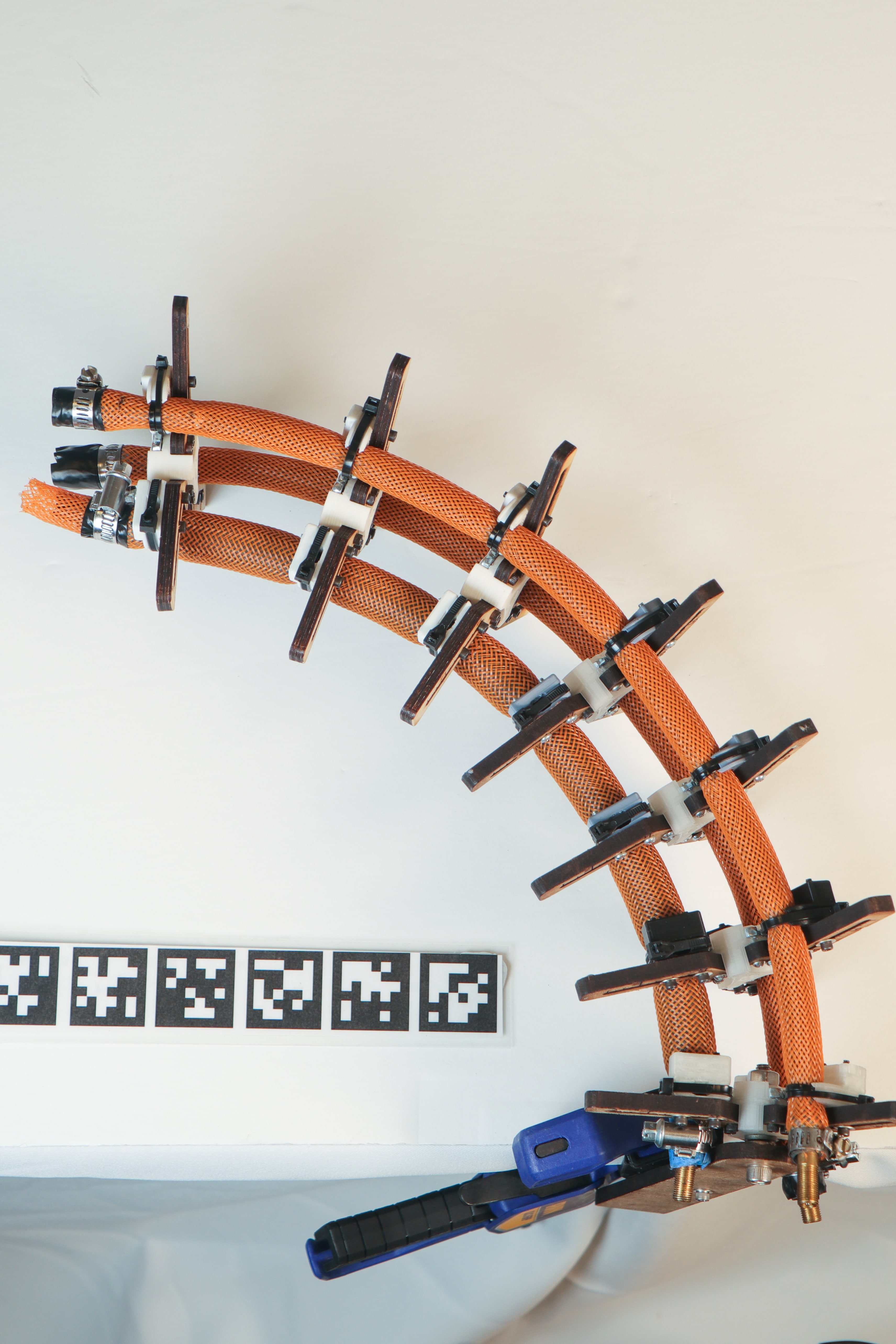}
        \centering
    \end{subfigure}
    \hfill
    \begin{subfigure}[t]{0.32\columnwidth}
        \includegraphics[width=\textwidth]{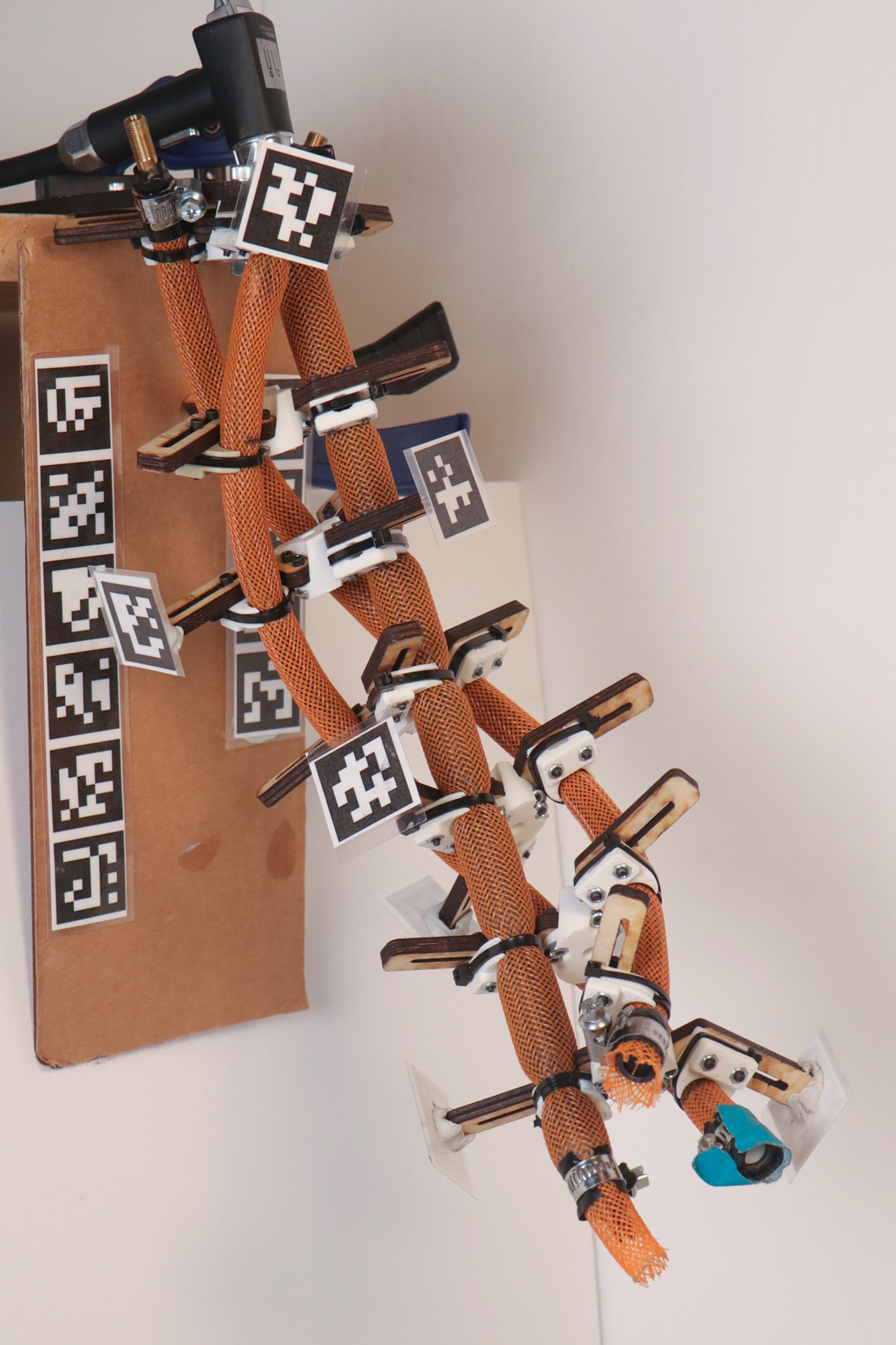}
        \centering
    \end{subfigure}
    
    \begin{subfigure}[t]{0.32\columnwidth}
        \includegraphics[width=\textwidth]{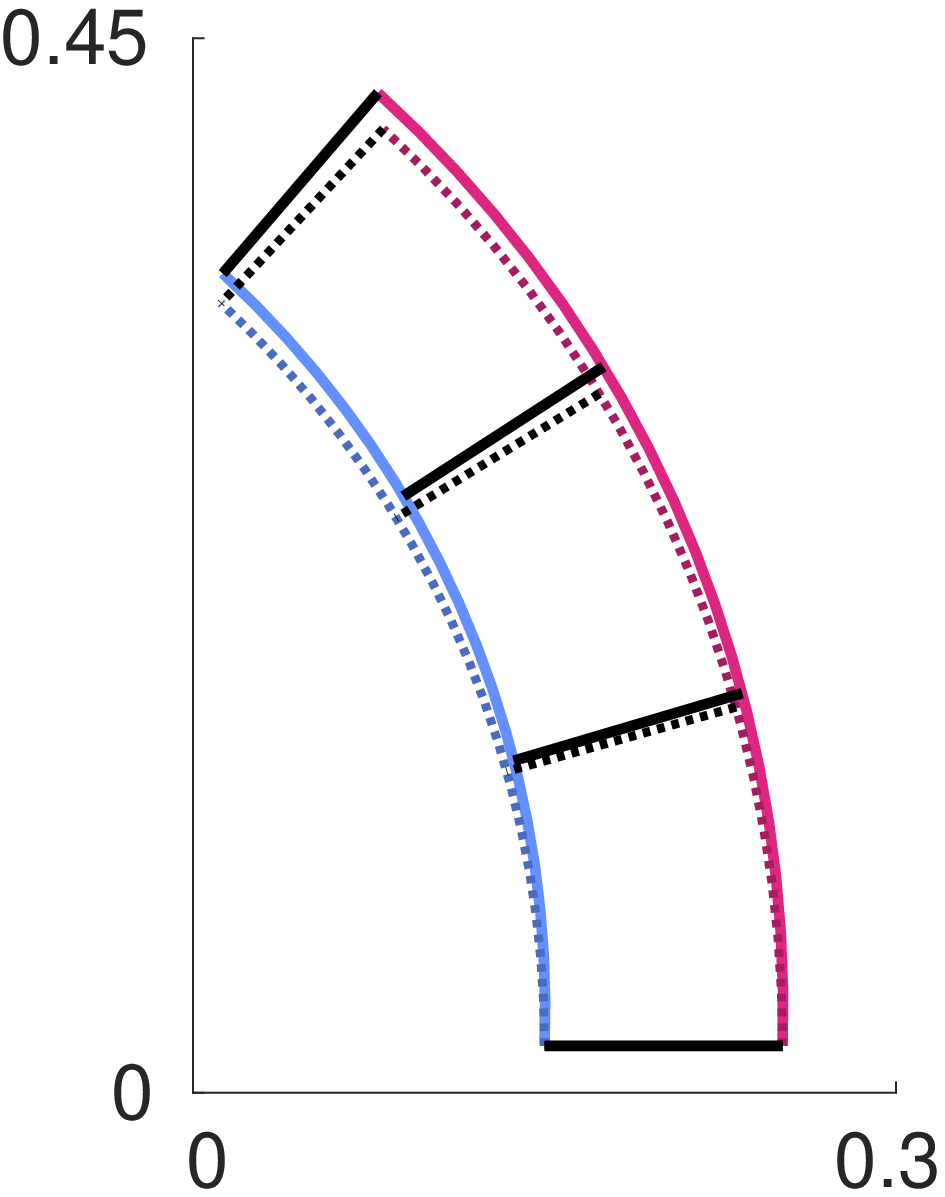}
        \centering
    \end{subfigure}
    \hfill
    \begin{subfigure}[t]{0.32\columnwidth}
        \includegraphics[width=\textwidth]{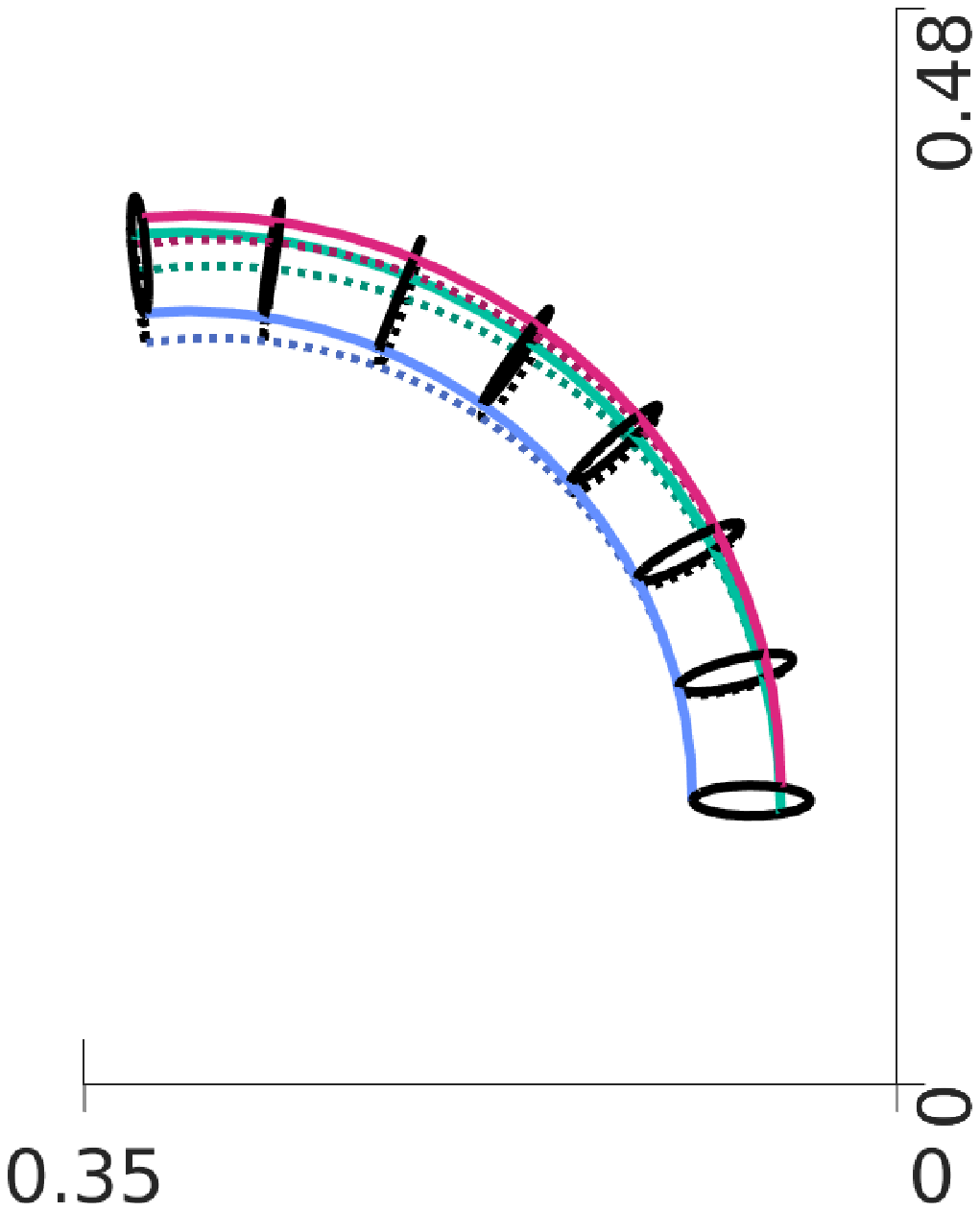}
        \centering
    \end{subfigure}
    \hfill
    \begin{subfigure}[t]{0.32\columnwidth}
        \includegraphics[width=\textwidth]{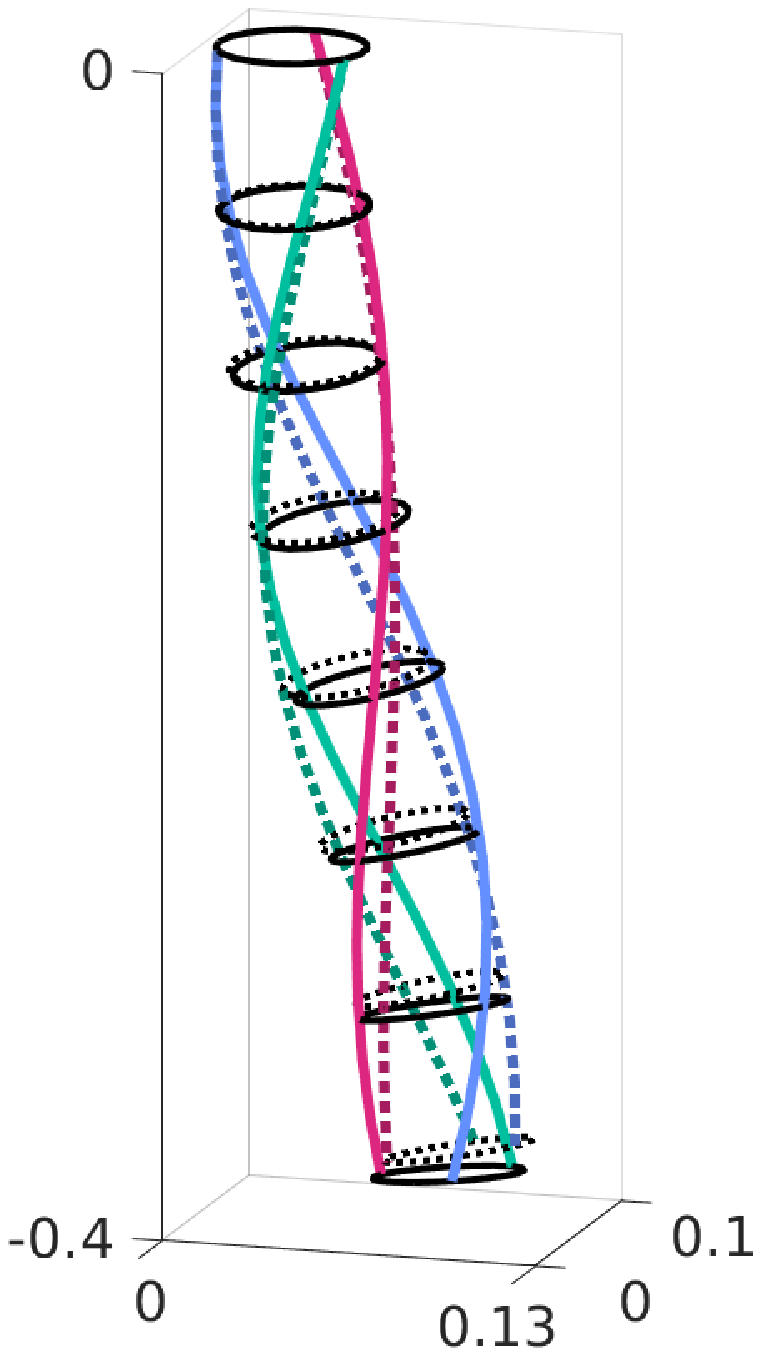}
        \centering
    \end{subfigure}
    
    \caption{Manipulators that exhibit constant bending and twisting. \textbf{Left to right}: Planar, spatial, and helical arms. \textbf{Top:} Physical systems with Apriltags attached for motion capture. \textbf{Bottom:} Model (dashed) versus measured (solid) geometry. All shown axes are in meters. The same model and parameters were used for all above experiments.}
    \label{fig:mckibben_arms}
\end{figure}

The design space of possible actuator routings and skeleton geometry is infinitely large.
Within this space, a common configuration consists of a few segments connected in series, where each section is composed of actuators that are arranged parallel to each other and the base-curve of the manipulator. The manipulator cross-section geometry is consistent throughout the length of each section (Fig \ref{fig:mckibben_arms}). 
The simple geometry of these manipulators makes their kinematics analytically tractable: using entirely geometric methods, the configuration of the actuators can be algebraically mapped to the length, curvature, and bending plane of the manipulator.
Prior work has developed models for use in control and motion planning, which has enabled applications in surgical robots, manipulation, and inspections \cite{Chen2022AOpportunities}. 

%
However, most bending continuum manipulators lack a torsional degree of freedom, and thus are incapable of separately controlling position and orientation, or performing motion primitives such as wrapping. 
%
Inspired by muscle structures in cephalopod tentacles, recent works \cite{Blumenschein2018HelicalBody}\cite{Starke2017OnRobots}\cite{Rozaidi2022AHISSbot} have shown that helically routing actuators at a constant pitch and diameter drastically enlarges the manipulator workspace by enabling constant twisting: a combination of bending and torsion.

This relatively simple design change invalidates constant curvature kinematics, thus requiring the use of the far more complex models such as Cosserat rod mechanics. \cite{Starke2017OnRobots}\cite{Rucker2011StaticsLoading}. The use of numerical integration makes Cosserat rod mechanics general and inclusive of non-constant deformation; however, it also makes them opaque. The reliance on integration to solve for manipulator deformation makes the approach computationally expensive for use in control and planning algorithms, which require repetitive kinematic solutions.

In this work, we derive an extension of constant curvature kinematics to include constant twist, while maintaining the simplicity of the approach. This kinematic model is purely algebraic and does not require numerical integration. Paired with an energy-based physics model, we produce a linear mapping between the actuator lengths and the deformed state of an unloaded manipulator. The model's form is analogous to traditional robot forward kinematics, and through our kinematic extension generalizes between constant curvature and constant twist manipulators.

This paper is structured as follows. Section \ref{section:prior_work} begins with an overview of kinematics, physics, and hybrid modeling approaches.
Section \ref{section:kinematics} reviews Cosserat-rod kinematics and derives the kinematic relation that unifies constant curvature kinematics to include constant twist.
Section \ref{section:equilibrium_model} derives our novel equilibrium model. %
Section \ref{section:results} presents results comparing the model to measured manipulator deformation from experiments.
Finally, we end with a discussion of results and considerations of future work in Section \ref{section:conclusion}.

\section{Prior Work}\label{section:prior_work}


The most well established class of continuum manipulator models are models that assume constant curvature \cite{Webster2010DesignReview}, which parameterize manipulator deformations by their length, curvature, and bending-plane. By assuming perfect knowledge of actuator lengths, the kinematics problem is reduced to pure geometry, thus enabling the development of models that, albeit nonlinear, function similarly to traditional robot forward kinematics. 

These models have proven useful for implementation in control and planning applications. However, they rely on an extrinsic parameterization of manipulator geometry. The notions of bending planes and tangent circles exist within a broader embedding space, and are specific to the structure of the manipulator. Thus, they do not generalize well to changes in manipulator design, such as additional actuators, or helically routed actuators. While recent works \cite{Blumenschein2018HelicalBody} have developed purely geometric kinematic models for constant twist manipulators, these models still parameterize the twisting geometry extrinsically as a winding radius and pitch. Therefore they suffer from the same lack of generality as models for constant curvature manipulators.


Cosserat rod kinematics present an alternative formulation of kinematics \cite{Rucker2011StaticsLoading}\cite{Renda2014DynamicCables}. By representing rod geometry intrinsically as the body-frame spatial velocity of a frame moving along the length of a rod, Cosserat rod kinematics is able to generalize to manipulators of arbitrary designs, including ones with non-constant deformation. While this requires numerical integration in the general case, restricting the model to constant deformation makes the kinematics analytically integrable. Thus, recent works have begun to apply this intrinsic parameterization to constant curvature kinematics \cite{Olson2020AnLoads} to derive more general models. The generalization of this parameterization to constant twist kinematics is natural, but has not yet been formulated in the literature. Thus, there does not currently exist a general form of constant twist kinematics.


A general constant deformation kinematics allows for the simplification of physics-based models. In \cite{Camarillo2008MechanicsManipulators}, the authors derived a linear mapping from tendon tensions to actuator deformation analogous to traditional forward kinematics. Combining physics-based methods with constant curvature kinematics lead to the development of a model applicable to manipulators with an arbitrary number of actuators while maintaining the simplicity of linearity. More recently, in \cite{Olson2020AnLoads} the authors instead focus on how manipulator behavior arises geometry between individual actuators. Similar to \cite{Camarillo2008MechanicsManipulators}, force-moment-balances are solved for within a manipulator with constant curvature kinematics. This time, by formulating the physics upon individual actuator mechanics the model generalizes across manipulators with different designs without the need for parameter re-fitting.

Due to the lack of a general constant twist kinematics, a similar work that applies physics-based modeling to constant twist manipulators does not yet exist. Thus, in this work we present the first generalized formulation of constant-twist kinematics, which we use to derive the first physics-based model applied specifically to constant-twist manipulators.


\section{ACTUATOR KINEMATICS}\label{section:kinematics}
We begin with a review of the Cosserat-rod parameterization of rod geometry to provide background and establish our notation. We then present a re-framing of constant curvature kinematics that leads to a straightforward derivation of a model of manipulator geometry at unloaded equilibrium.

\subsection{Individual Rod Kinematics}
In 3D Cosserat-rod theory, a rod is characterized by its center-line curve in 3D space $\bm{p}(s) \in \mathbb{R}^3$ and its material orientation $R(s) \in SO(3)$. The reference parameter $s \in [0, 1]$ denotes the location of the current position along the length of the arm; ie, $\bm{p}(0)$ is the position at the base of the rod while $p(1)$ the position at the tip.

The positions and orientations of frames along the rod can be combined into a single 6-DOF pose $ g(s)$, and element of the Lie group $SE(3)$, whose matrix representation is a homogeneous transformation matrix. We denote the matrix representation of a Lie group or Lie algebra element with $\rho(\cdot)$:

\begin{equation}
    \rho(g(s)) = 
    \begin{bmatrix} 
        R(s) & \bm{p}(s) \\
        \mathbf{0} & 1
    \end{bmatrix}
\end{equation}

Alongside the poses $g(s)$, we consider its rate of change along the length of the rod:

\begin{equation}\label{eqn:g_gcirc}
    \frac{\partial}{\partial s} \rho(g) = \rho(\dot{g}) = \rho(g) \rho(\rightgroupderiv{g})
\end{equation}

where the \textit{twist-vector} $\rightgroupderiv{g}(s)$ is an element of the Lie algebra $\mathfrak{se}(3)$ - a vector $\begin{bmatrix}\bm{v} & \bm{\omega} \end{bmatrix}^T$ that contains the body-frame linear and angular velocities $\bm{v}$ and $\bm{\omega}$. The matrix representation is:

\begin{equation}
\rho(\rightgroupderiv{g}(s)) = 
    \begin{bmatrix}
        \rho(\bm{\omega}(s)) & \bm{v}(s) \\
        \mathbf{0} & 0
    \end{bmatrix}
\end{equation}

where $\rho(\bm{\omega})$ is the skew-symmetric matrix representation of an angular velocity in $\mathfrak{so}(3)$.

The rate of change $\rightgroupderiv{g}(s)$ alone can parameterize a rod's poses $g(s)$ along the entire rod for all $s$. In general, body-frame twists $\rightgroupderiv{g}(s)$ along the length of the rod can be converted to world-frame and numerically integrated to recover the rod geometry. In the special case where the $\rightgroupderiv{g}(s) = \rightgroupderiv{g}$ is constant over $s$, this integration can be done analytically through exponentiation.

\begin{equation}\label{eqn:g_exp}
    g(s) = g(0) \circ \exp(s \ \rightgroupderiv{g})
\end{equation}

Or, in matrix representation, with $\exp_M$ as the standard matrix exponential:

\begin{equation}
    \rho(g(s)) = \rho(g(0)) \circ \exp_M(s \ \rho(\rightgroupderiv{g}))
\end{equation}

\begin{figure}
    \centering
    \includegraphics[width=\columnwidth]{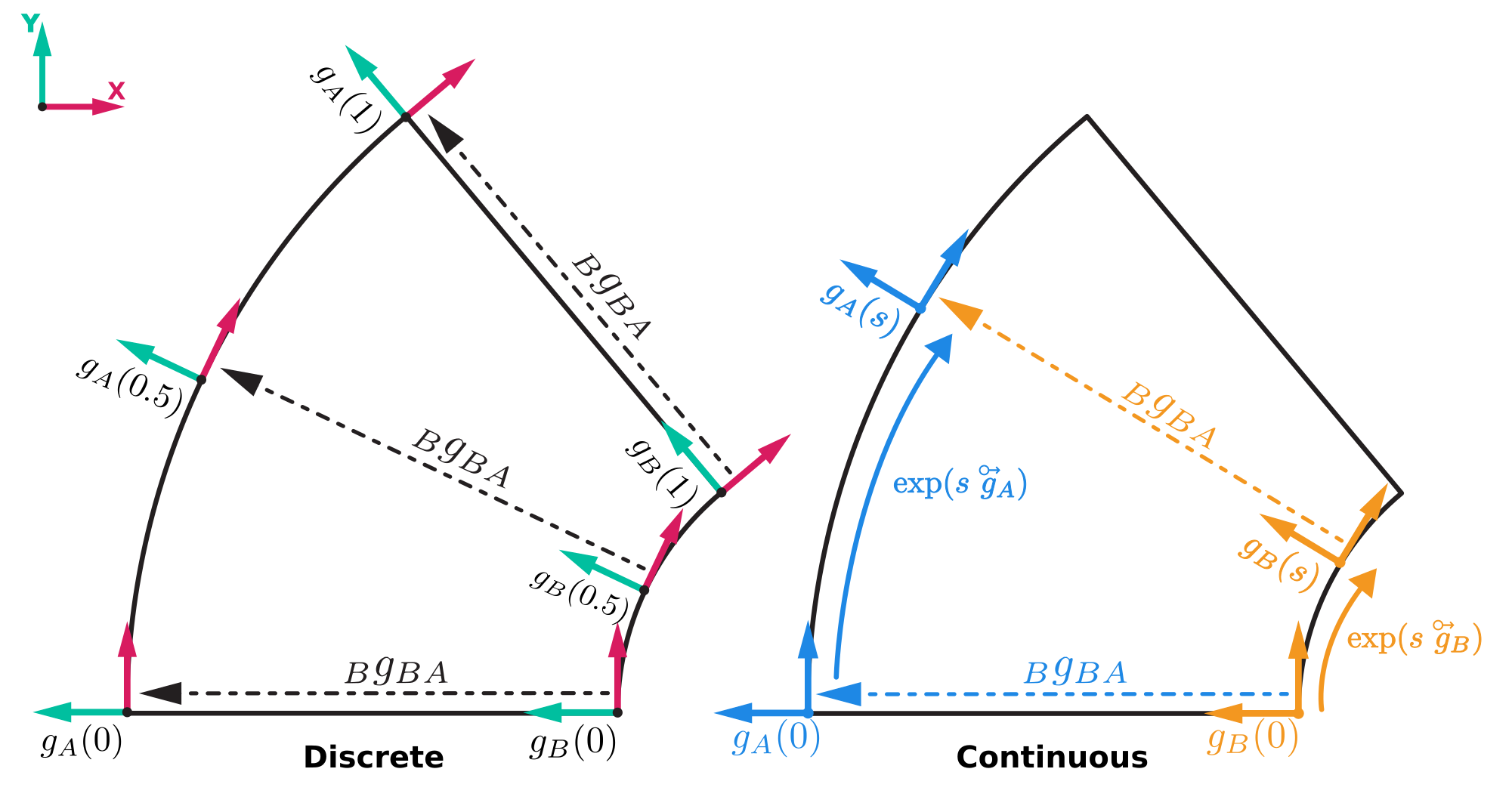}
    \caption{Let $g_A(\cdot)$ and $g_B(\cdot)$ refer to poses along actuators $A$ and $B$. Assuming constant curvature means assuming the cross-section geometry ${}_Bg_{BA}$ is uniform throughout the arm. This enables relating the twist-vectors of different actuators $\rightgroupderiv{g}_A$ and $\rightgroupderiv{g}_B$ as the velocities of segments on different paths (blue and yellow) that lead to the same destination.}
    \label{fig:planar_arm_uniformity_assumption}
\end{figure}

As an example, consider the twist-vector $\rightgroupderiv{g}_0$ of an actuator in its neutral configuration. In its neutral state, we assume an actuator takes on its free-contraction length with zero shearing, torsion, or bending. This free-contraction length $l(q)$ can be experimentally measured as a function of the actuator input $q$, such as muscle pressure or tendon pulley angle. The actuator's neutral twist-vector $\rightgroupderiv{g}_0$ is thus $\begin{bmatrix}l(q) & 0 & 0 & 0 & 0 & 0 \end{bmatrix}^T$. Integrating this quantity along the unit interval $[0, 1]$ recovers the positions along the undeformed rod.

Any rod with a constant twist-vector $\rightgroupderiv{g}$ that deviates from this neutral configuration $\rightgroupderiv{g}_0$ must therefore entail deformation. Taking the difference $\rightgroupderiv{g} - \rightgroupderiv{g}_0$ gives us the deformation vector $\Delta \rightgroupderiv{g} = \begin{bmatrix} \lambda & \gamma_y & \gamma_z & \tau & \omega_y & \omega_z \end{bmatrix}^T$, which respectively contains the stretch, y- and z-shearing, x-curvature (torsion)-, and y- and z- curvature. We will later combine this deformation vector with physical laws to calculate quantities such as force and elastic energy.

Having covered both the twist-vector parameterization of elastic rod geometry and deformation, we conclude our review of Cosserat rod kinematics theory for individual rods.

\subsection{Relating Actuator Twists}

For a constant-twist manipulator we derive a novel linear mapping from the twist-vectors of each actuators to the twist-vector of the manipulator base-curve. We do so by generalizing the geometric constraints between the actuators.

It is established that for manipulators assumed to be constant-curvature, the uniformity of the cross-section geometry is crucial to the validity of the constant-curvature assumption \cite{Li2002DesignRobots}. Works on constant-twist manipulators similarly rely on designs with a uniform helical winding pitch \cite{Blumenschein2018HelicalBody}. In designs where actuators are attached to a backbone with consistent cross-sections, this uniformity is inherited from the backbone. Other designs lack a continuous backbone and employ a set of inter-actuator separators: enough separators are used so that constraints between actuators are still enforced, even at points not directly constrained by separators, as in Fig. \ref{fig:planar_arm_uniformity_assumption}. For either design, the result is the same: a uniform cross-section geometry leads to the constant deformation throughout the manipulator. 

Note that this is only true in the un-loaded case: the addition of external loads causes deformation that is no longer uniform throughout. 


Consider a planar arm with two actuators $A$ and $B$, as depicted in Fig. \ref{fig:planar_arm_uniformity_assumption}. We can represent the cross-section geometry as the transformation from a frame attached to actuator $B$ to a frame attached to actuator $A$:

\begin{equation} \label{eqn:g_0}
    g_A(s) = g_B(s) {}_Bg_{BA}
\end{equation}

where $_{B}g_{BA}$ is an element of $SE(3)$ representing the transformation from a frame on actuator $B$ to a frame on actuator $A$, with respect to the frame on $B$. Note that due to the uniform geometry, this transformation is constant across the entire manipulator. We can exploit this fact to relate the twist-vectors between different actuators.

Expressing the poses $g(s)$ along each actuator by exponentiating the twist-vectors using Eqn. \ref{eqn:g_exp} gives:

\begin{equation}
    g_A(0) \circ \exp(s \ \rightgroupderiv{g}_A) = g_B(0) \circ \exp(s \ \rightgroupderiv{g}_B) \circ  {}_Bg_{BA}
\end{equation}

Eqn. \ref{eqn:g_0} gives us $g_A(0)^{-1}g_B(0) = {}_Bg_{BA}^{-1}$, which we substitute in:

\begin{equation}
    \exp(s \ \rightgroupderiv{g}_A) =  {}_Bg_{BA}^{-1} \circ \exp(s \ \rightgroupderiv{g}_B) \circ {}_Bg_{BA}
\end{equation}

We have thus arrived at the common definition for the Adjoint matrix. Thus the twist-vectors $\rightgroupderiv{g}_A$ and $\rightgroupderiv{g}_B$ are related via the Adjoint matrix in $SE(3)$:

\begin{equation}\label{eqn:adjoint_inverse_for_BA}
    \rightgroupderiv{g_A} = \Adjointinv{BA} \rightgroupderiv{g}_B
\end{equation}

The structure of the Adjoint matrix $\Adjoint{BA}$ depends on the Lie group used to parameterize actuator geometry, such as $SE(2)$ or $SE(3)$. The values of the matrix depend on the cross-section geometry: ${}_Bg_{BA}$. For a full derivation of the Adjoint map and matrix representations in common groups we refer the reader to \cite{Sola2018ARobotics}.
3
Note that so long as the cross-section geometry is uniform, one can apply equation \ref{eqn:adjoint_inverse_for_BA} to any imagined actuator given its location in the manipulator cross-section. For a general manipulator with an arbitrary number of actuators, we therefore relate all actuators to a single reference actuator. If the reference actuator is considered at the center of the manipulator, it traces the base-curve of the manipulator. Thus, for any actuator $i$ with transformation ${}_ig_{oi}$ relative to the base-curve, its twist-vector $\rightgroupderiv{g}_i$ can be related to the base-curve twist-vector $\rightgroupderiv{g}_o$ by the Adjoint matrix $\Adjoint{oi}$

\begin{equation}\label{eqn:adjoint_inverse_for_each_muscle}
    \rightgroupderiv{g}_i = \Adjointinv{oi} \rightgroupderiv{g}_o
\end{equation}

We have thus shown that assuming the cross-section geometry is uniform throughout the continuum manipulator, the twist-vector of every actuator in the manipulator can be related to the twist-vector of the manipulator base-curve via the Adjoint map. This concludes the derivation of our kinematic model.

\section{Equilibrium Mechanics Model} \label{section:equilibrium_model}
We now derive our equilibrium manipulator model by formulating the elastic energy stored within each actuator, and solving for the manipulator configuration that minimizes the total energy. Note once again that this model does not consider external loading.

Combining the deformation vector $\Delta \rightgroupderiv{g}$ from before with a Hookean spring law with stiffness matrix $K = \textrm{diag}(\begin{bmatrix} K_\epsilon & K_\gamma & K_\gamma & K_\tau & K_\kappa & K_\kappa\end{bmatrix})$, we arrive at the standard expression for local wrench moments throughout the length of the rod:

\begin{equation}
    \bm{f} = K (\rightgroupderiv{g} - \rightgroupderiv{g}_0)
\end{equation}

Note that by being a spatial-velocity, the linear component $\bm{v}$ of $\rightgroupderiv{g}$ has units [m/m] and is thus unitless, while the angular component $\bm{\omega}$ has units [rad/m]. Thus the linear stiffness constants $K_\epsilon$ and $K_\gamma$ are have units of [N], and the rotational stiffness constants $K_\tau$ and $K_\kappa$ have units of  [N m\textsuperscript{2}], because radians are unitless.

One can also integrate these local wrenches to find the elastic energy stored within a rod:
\begin{align}\label{eqn:single_beam_energy}
    U(q) &= \int_{\rightgroupderiv{g}_0}^{\rightgroupderiv{g}} K \bm{r} \cdot d\bm{r} \\
    &= \frac{1}{2}(\rightgroupderiv{g} - \rightgroupderiv{g}_0)^T K (\rightgroupderiv{g} - \rightgroupderiv{g}_0)
\end{align}

Remember that $\rightgroupderiv{g}_0$ is dependent on the actuator configuration $q$.
We will now use the Adjoint mapping between actuator twist-vectors to derive our equilibrium model.

The total elastic energy stored within a manipulator is the sum of the elastic energy stored in each actuator. For a manipulator with $n$ actuators, using Eqn. \ref{eqn:single_beam_energy} we can express the total energy as:

\begin{equation}
    U_T = \sum_{i=1}^n U_i = \frac{1}{2} \sum_{i=1}^n (\rightgroupderiv{g}_i - \rightgroupderiv{g}_{0_i})^T K (\rightgroupderiv{g}_i - \rightgroupderiv{g}_{0_i})
\end{equation}

Using Eqn. \ref{eqn:adjoint_inverse_for_each_muscle}, we can then represent each actuator's twist-vector $\rightgroupderiv{g}_i$ in terms of the base-curve twist-vector:

\begin{align*}
    U_T &= \frac{1}{2} \sum_{i=1}^n(\Adjointinv{oi}\rightgroupderiv{g}_o - \rightgroupderiv{g}_{0_i})^T K (\Adjointinv{oi} \rightgroupderiv{g}_o  - \rightgroupderiv{g}_{0_i})
\end{align*}

Expanding the quadratics gives:
\begin{align*}
    U_T = \frac{1}{2}\sum_{i = 1}^n( \rightgroupderiv{g}_o^T \Adjointinvtpose{oi} K \Adjointinv{oi} \rightgroupderiv{g}_o
    - 2\rightgroupderiv{g}_o^T \Adjointinvtpose{oi} K\rightgroupderiv{g}_{0_i} + \rightgroupderiv{g}_{0_i}^TK\rightgroupderiv{g}_{0_i})
\end{align*}

Recall that the reference twist-vector $\rightgroupderiv{g}_{0_i}$ for each rod is a function of its configuration $q_i$, where $\rightgroupderiv{g}_{0_i} = \begin{bmatrix} l(q_i) & 0 & 0 & 0 & 0 & 0\end{bmatrix}^T = l(q_i) \mathbf{e_1}$ and $\mathbf{e_1}$ is the first basis vector. The equilibrium base-curve twist-vector $\rightgroupderiv{g}_o^*$ is the collective actuator configuration where the gradient of the elastic energy is the zero vector. Taking the gradient, setting it to zero, and solving for $\rightgroupderiv{g}_o^*$ gives the following:

\begin{align*}
    \nabla U_T &=  \sum_{i = 1}^n (\Adjointinvtpose{oi} K \Adjointinv{oi} \rightgroupderiv{g}_o^* - \Adjointinvtpose{oi} K\rightgroupderiv{g}_{0_i}) = \mathbf{0} \\
    &\Rightarrow \sum_{i = 1}^n \Adjointinvtpose{oi} K \Adjointinv{oi}\rightgroupderiv{g}_o^* = \sum_{i=1}^n \Adjointinvtpose{oi} K \mathbf{e_1} l(q_i)
\end{align*}

The $\rightgroupderiv{g}_o^*$ in the LHS of the above equation can be factored out of the sum. The RHS equation can be simplified by noting that it is a linear combination with each $l(q_i)$ element. Combining each $l(q_i)$ element into a vector gives the following:

\begin{multline}\label{eqn:full_factored}
    \sum_{i = 1}^n (\Adjointinvtpose{oi} K \Adjointinv{oi}) \rightgroupderiv{g}_o^*
    = \\
    \begin{bmatrix}
        \Adjointinvtpose{o1} K \mathbf{e_1} &
        \Adjointinvtpose{o2} K \mathbf{e_1} &
        \dots &
        \Adjointinvtpose{on} K \mathbf{e_1}
    \end{bmatrix}
    \begin{bmatrix}
        l(q_1) \\
        l(q_2) \\
        \vdots \\
        l(q_n)
    \end{bmatrix}
\end{multline}

Using $\mathbf{A} = \sum_{i = 1}^n (\Adjointinvtpose{oi} K \Adjointinv{oi})$ as the sum of the Adjoints scaled with elastic constants, the matrix $\mathbf{D} = \begin{bmatrix}
        \Adjointinvtpose{o1} K \mathbf{e_1} &
        \Adjointinvtpose{o2} K \mathbf{e_1} &
        \dots &
        \Adjointinvtpose{on} K \mathbf{e_1}
\end{bmatrix}$, and vector $\mathbf{l} =     \begin{bmatrix} l(q_1) & l(q_2) & \dots & l(q_n)
\end{bmatrix}^T$ as the vector of actuator neutral lengths dependent on pressure, we can represent Eqn. \ref{eqn:full_factored} in compact matrix form:

\begin{equation} \label{model_final}
    \mathbf{A}\rightgroupderiv{g}_o^* = \mathbf{D l}
\end{equation}

As a final step, the neutral configuration can be solved for by taking the pseudo-inverse of $\mathbf{A}$:

\begin{equation}\label{eqn:model_final}
    \rightgroupderiv{g}_o^* = \mathbf{A^+Dl}
\end{equation}

This concludes our derivation of our simplified rod model for the equilibrium configuration of an unloaded constant twist manipulator.

\subsection{Example In Coordinates}
We now provide the coordinate representation of our model of a two-actuator planar arm with width $2d$. Since the planar arm exists entirely in two dimensions, we embed the model in $SE(2)$. By using the $SE(2)$ twist-vector form $\rightgroupderiv{g} = \begin{bmatrix} \lambda & \gamma & \kappa \end{bmatrix}$ respectively containing length, shear, and curvature, and the $SE(2)$ Adjoint as derived in \cite{Sola2018ARobotics}, Eqn. \ref{eqn:model_final} takes on the following form:

\begin{equation}\label{eqn:planar_2muscle}
    \begin{bmatrix}
        2K_\epsilon & 0 & 0 \\
        0 & 2K_\gamma & 0 \\
        0 & 0 & 2K_\kappa + 2K_\epsilon d^2
    \end{bmatrix}
    \begin{bmatrix}
        \lambda \\ \gamma \\ \omega
    \end{bmatrix}
     = K_{\epsilon}
    \begin{bmatrix}
        1 & 1\\
        0 & 0\\
        -d & -d
    \end{bmatrix}
    \begin{bmatrix}
        l(q_1) \\
        l(q_2)
    \end{bmatrix}
\end{equation}

Because $K_\epsilon l(q_i)$ is proportional to the stress experienced in each actuator, our model is equivalent to that of \cite{Camarillo2008MechanicsManipulators}. However, our model generalizes to different manipulator designs, and crucially is able to predict how the overall manipulator bending stiffness $2K_\kappa + 2K_\epsilon d^2$ changes with width - something that was left experimentally determined in \cite{Camarillo2008MechanicsManipulators}. As that work was performed on cable-driven manipulators while ours is with pneumatic ones, our model generalizes to manipulators with different types of actuation.

\section{EXPERIMENT RESULTS}\label{section:results}
We now describe the setup, methodology, and results of validating our constant-twist equilibrium model on planar, spatial, and helical arms. We first describe the construction of our three manipulators with adjustable dimensions. We then outline the procedure of each experiment, as well as the parameter characterization process. Finally, we present the experiment results, concluded by a discussion of the sources of error.

\subsection{Manipulator Construction}
Individual actuators were constructed identically to \cite{Rozaidi2022AHISSbot}, but with an effective length of 460mm. These actuators were mounted to three different types of actuator-separators to form our array of manipulators.  Constraining a constant distance between two muscles creates a planar arm capable of purely planar bending. A spatial arm which can bend in three dimensions is created by adding a third muscle to form a radial arrangement. Finally, tilting the actuator mount relative to the separator creates a pitch, and thus a helical arm that both bends and twists. Pictures of the three actuators can be found in Fig. \ref{fig:mckibben_arms}.

For all manipulators, the muscles were held in place by clamps which constrained their position but allowed them to rotate in place. These clamps were then rigidly attached to muscle separators with adjustable mounting positions, enabling the effective width or diameter of the arms to be easily reconfigured. The planar arm was tested with widths of 50.8mm, 76.2mm, and 101.6mm widths. The spatial arm was tested with diameters of 50.8mm and 101.6mm. The helical arm was only tested at a diameter of 50.8mm.

\subsection{Experiment Procedure}

Laminated Apriltag \cite{Wang2016AprilTagDetection} fiducial markers were first attached to the manipulators using adhesive putty to enable motion capture. The experiments were then conducted as a series of static measurements. Using a bicycle pump, the manipulator was inflated to a desired pressure, which was maintained constant by an inline pressure regulator. Then, with a Logitech c920 webcam, we capture video footage of the fiducials which we later localize using TagSLAM \cite{Pfrommer2019TagSLAM:Markers}. Finally, the twist-vector $\rightgroupderiv{g}_o$ of the manipulator base-curve is fitted based on the measured tag positions through a nonlinear optimization solved with Matlab’s \verb!fminsearch! function. Between three to five such captures are performed for each muscle inflation pressure, and their median is taken as the final result.

\subsection{System Characterization}

\begin{figure}
    \centering
    \includegraphics[width=0.98\columnwidth]{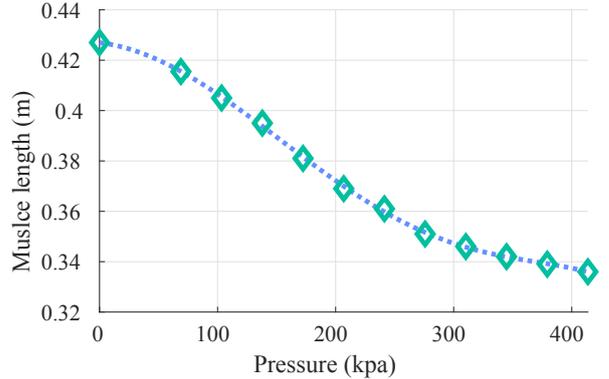}
    \caption{Fitted artificial muscle free-contraction line.}
    \label{fig:contraction_fit}
\end{figure}

Our equilibrium model is characterized by two sets of parameters. The first is the mapping from muscle inflation pressure to neutral length $l(q)$, which we refer to as the free-contraction model. The second parameter is the individual muscle stiffness matrix $K$. 

The contraction model was experimentally measured by inflating a single unladen muscle to various pressures within our activation range and measuring its length. As in \cite{Olson2020AnLoads}, we then fit a fifth-order polynomial over the measured values to arrive at our fitted model. The results can be seen in Figure \ref{fig:contraction_fit}a.

For the muscle stiffness matrix $K$, appropriate assumptions help simplify the full matrix to a single value: the ratio of strain stiffness $K_\epsilon$ to bending stiffness $K_\kappa$. First, because the muscles are not actually rigidly clamped to the separators and instead are allowed to rotate in place, we assume the effective torsional stiffness $K_\tau$ of a single muscle is 0. Second, we assume the muscles experience zero shearing, and thus have infinite shearing stiffness $K_\gamma$; in practice, we've found that using any shearing stiffness $K_\gamma$ more than ten times the strain stiffness $K_\epsilon$ had the same effect. Finally, we note that in equation \ref{eqn:full_factored} any scale factor of K would cancel each other out; we can therefore eliminate $K_\epsilon$ by normalizing K with $K_\epsilon$, and just consider the normalized bending stiffness $\bar{K_\kappa} = K_\kappa / K_\epsilon$. To recap, our final stiffness matrix is therefore $K = \textrm{diag}(1,\ 10,\ 10,\ 0,\ 0,\ \bar{K_\kappa})$. Recall the units of $K_\epsilon$ and $K_\gamma$ are [N], and the units of $K_\tau$ and $K_\kappa$ are [Nm$^2$].

In practice, our value for $\bar{K_\kappa}$ is taken as the value that minimizes tip position error for a single calibration dataset. This value is then used for all other datasets.

\subsection{Model Accuracy}

\begin{figure*}
    \centering
    \begin{subfigure}[b]{\columnwidth}
        \includegraphics[width=\textwidth]{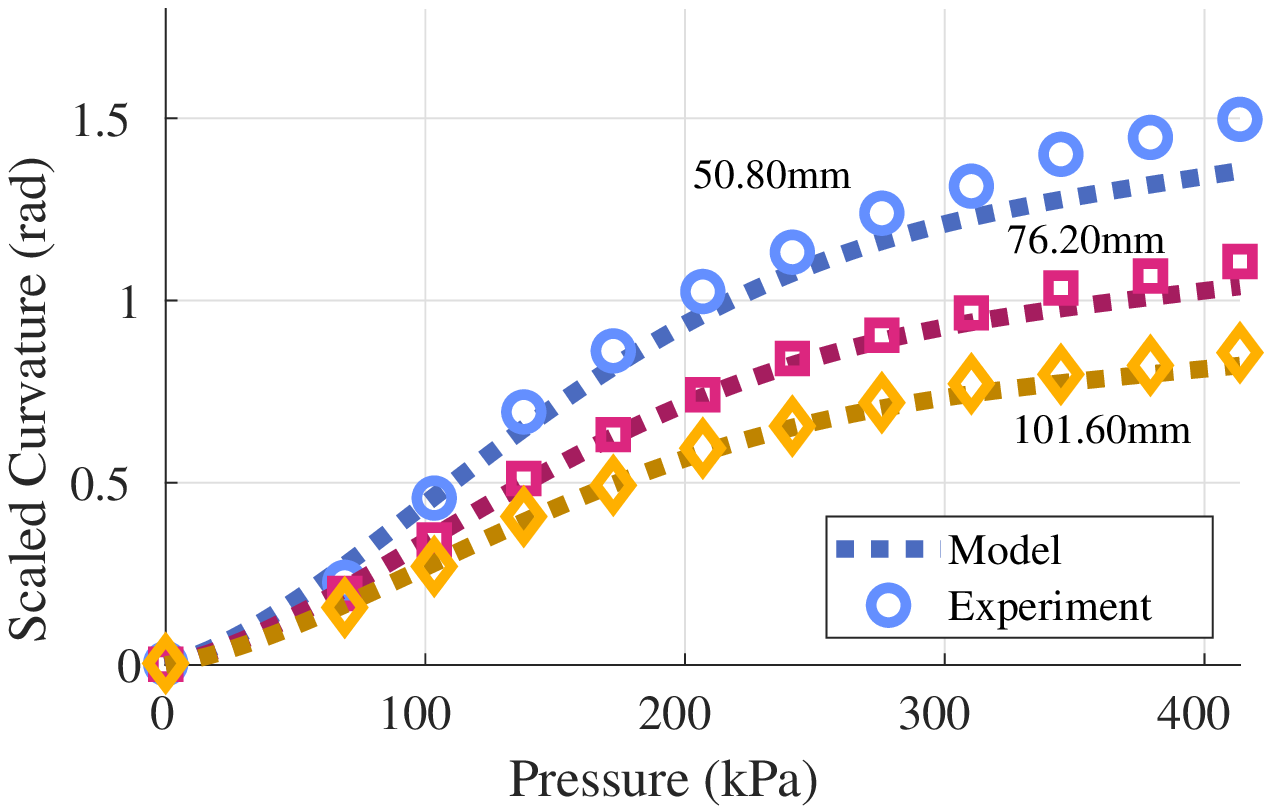}
        \centering
    \end{subfigure}
    \begin{subfigure}[b]{\columnwidth}
        \includegraphics[width=\textwidth]{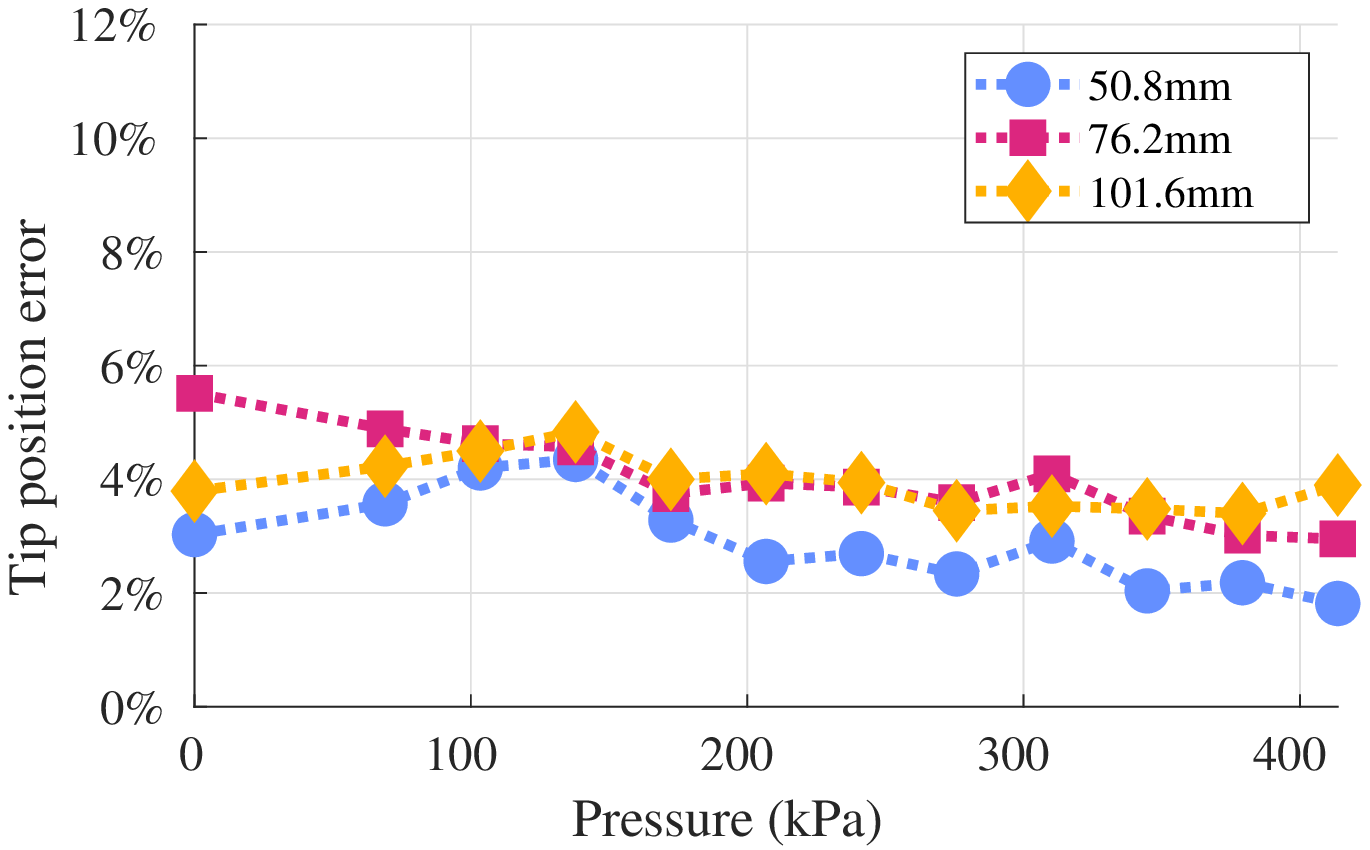}
        \centering
    \end{subfigure}
    
    \begin{subfigure}[b]{\columnwidth}
        \includegraphics[width=\textwidth]{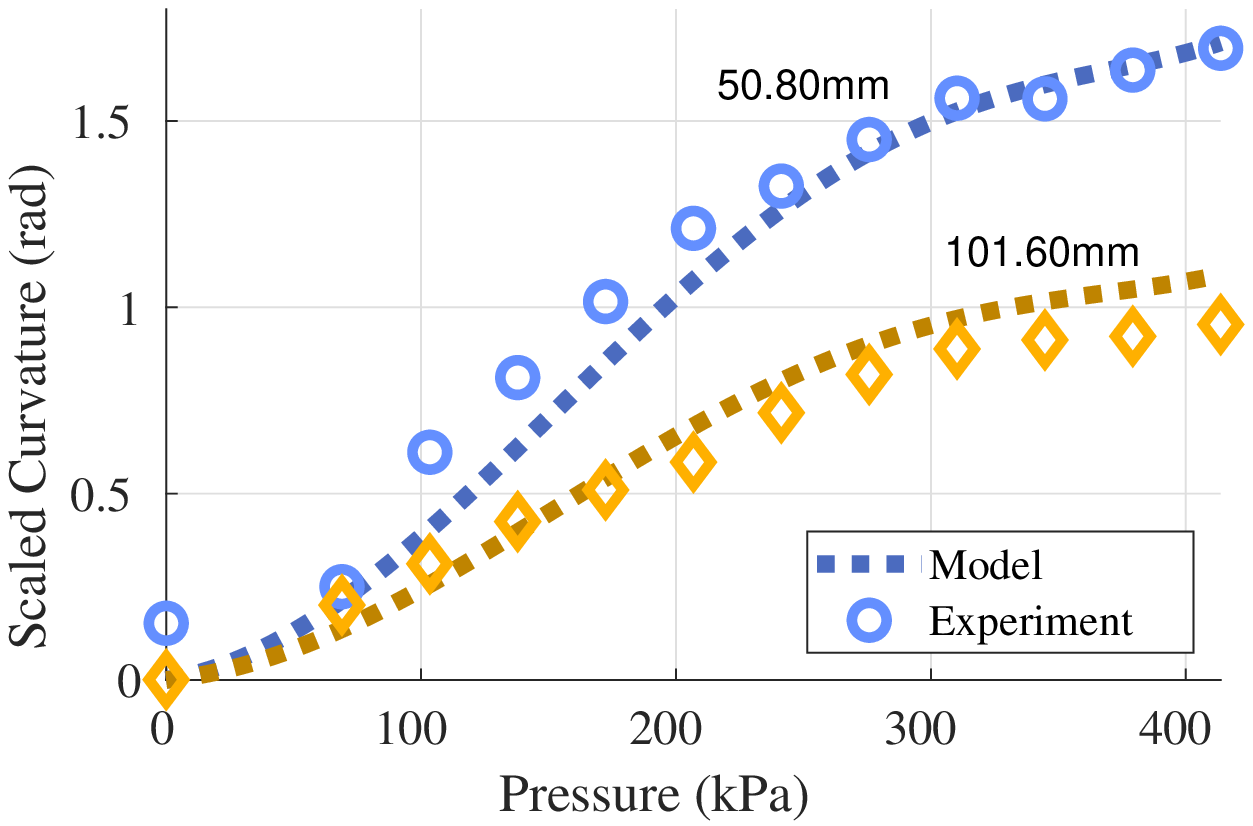}
        \centering
    \end{subfigure}
    \begin{subfigure}[b]{\columnwidth}
        \includegraphics[width=\textwidth]{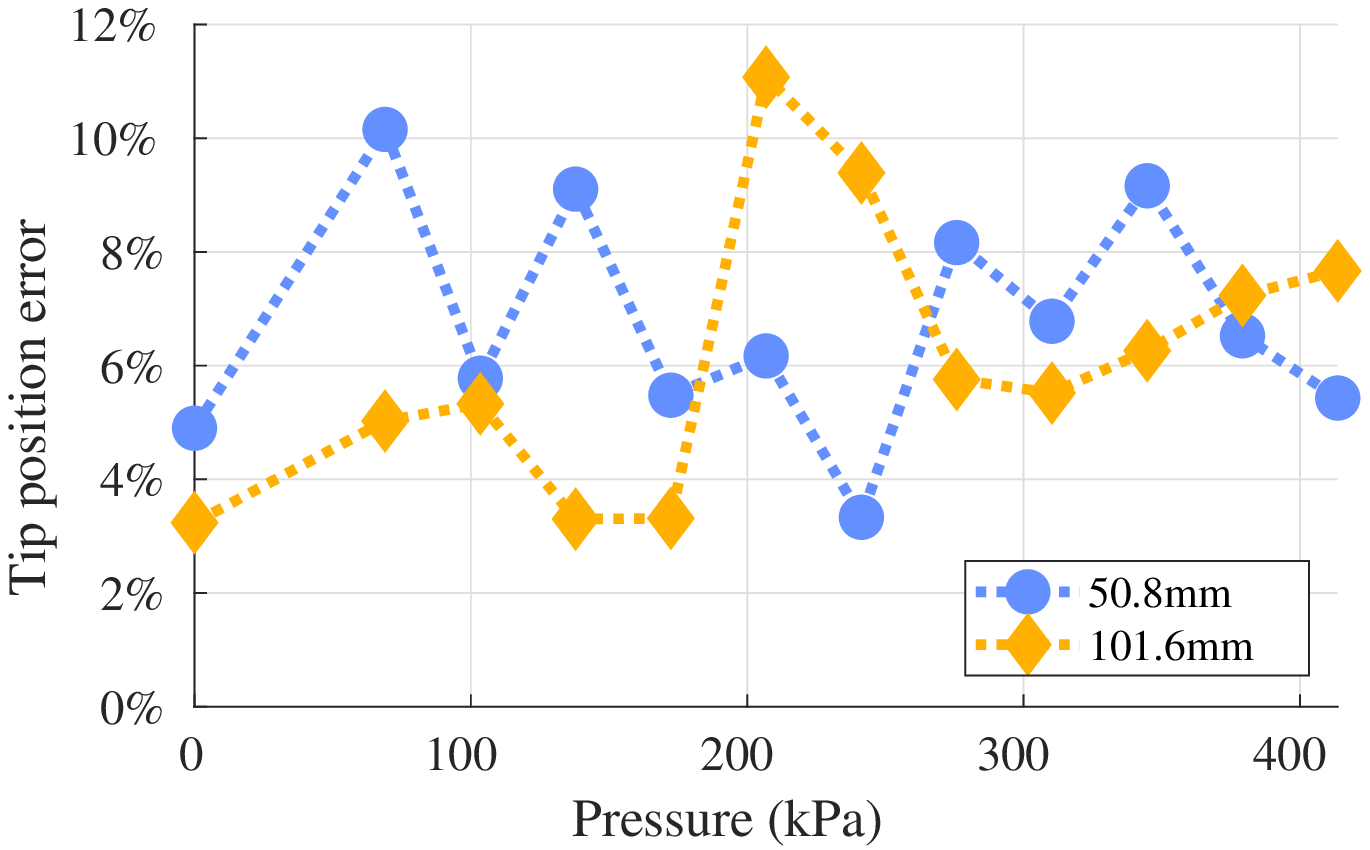} 
        \centering
    \end{subfigure}

    \caption{\textbf{Planar (}Top\textbf{), and Spatial (}Bottom\textbf{) arm model accuracy for a range of arm diameters.} Left: Model vs. experimentally measured length-scaled curvature, which for a 2D arm is equivalent to the bending angle. Right: Manipulator tip position error normalized by the length of the manipulator.}
    \label{fig:planar_spatial_arm_results}
\end{figure*}

We reiterate that the model in all following tests are parameterized by the same stiffness matrix $K$ and contraction model $l(q)$.

For each type of manipulator tested, we provide two metrics. The first is the accuracy of the arc-length-scaled-curvature over pressure, which we directly get through the corresponding component in the base-cure twist-vector. The authors of \cite{Olson2020AnLoads} used the same metric but with the true-curvature. The second is the tip position error: the difference between predicted and measured positions of the manipulator tips, normalized by the actuator's length. This metric is used in \cite{Rucker2011StaticsLoading}. 

Figures \ref{fig:planar_spatial_arm_results} shows the experiment results of a two-muscle planar and a three-muscle spatial arm at all dimensions tested. For each arm a single muscle was actuated to induce a planar bending. The bending plane was oriented such that the arms were evenly supported by a flat tabletop at all times, eliminating the effects of gravity.

\begin{figure}
    \centering
    \begin{subfigure}[b]{0.32\columnwidth}
        \includegraphics[width=\textwidth]{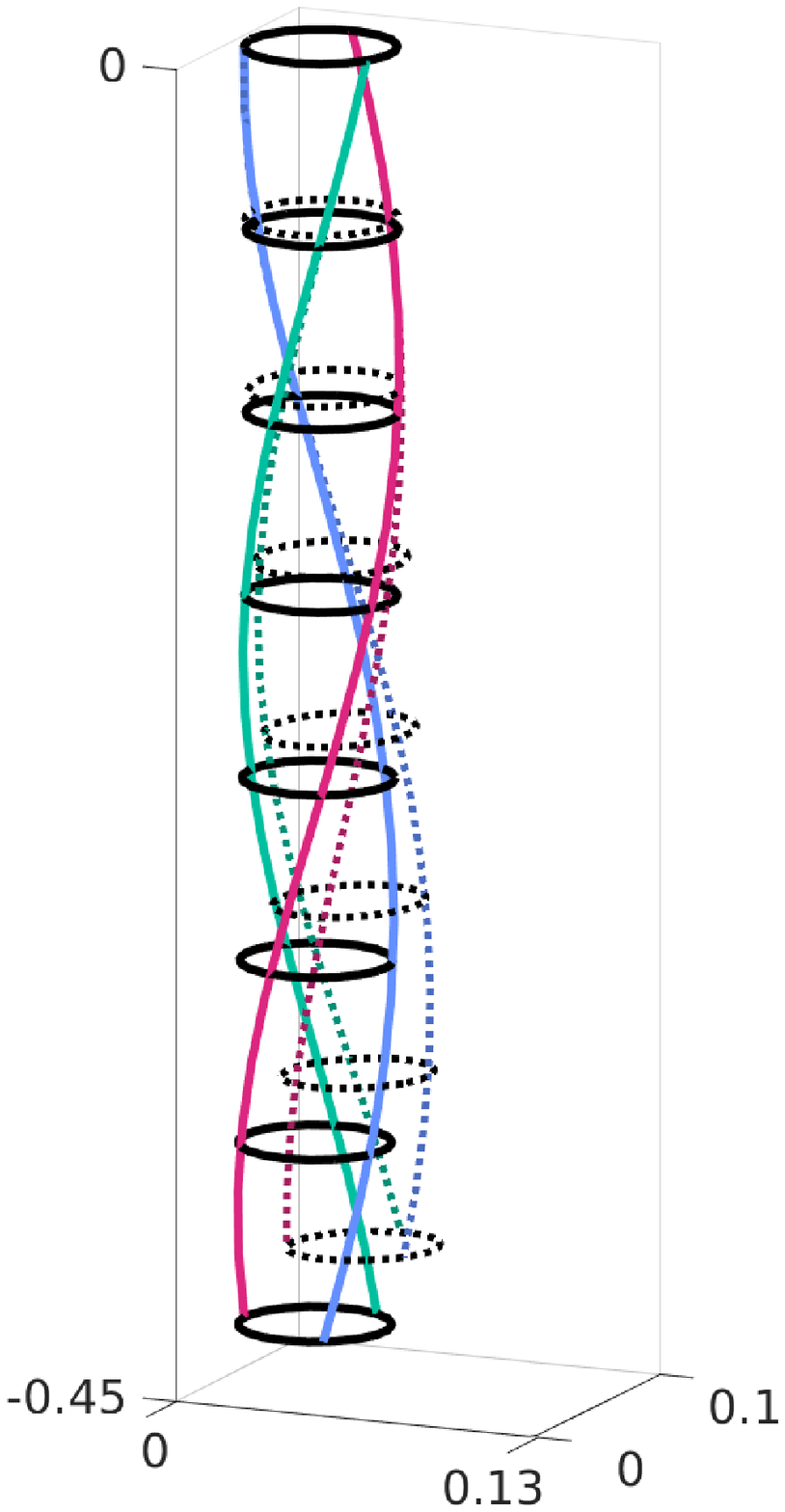}
        \centering
    \end{subfigure}
    \begin{subfigure}[b]{0.32\columnwidth}
        \includegraphics[width=\textwidth]{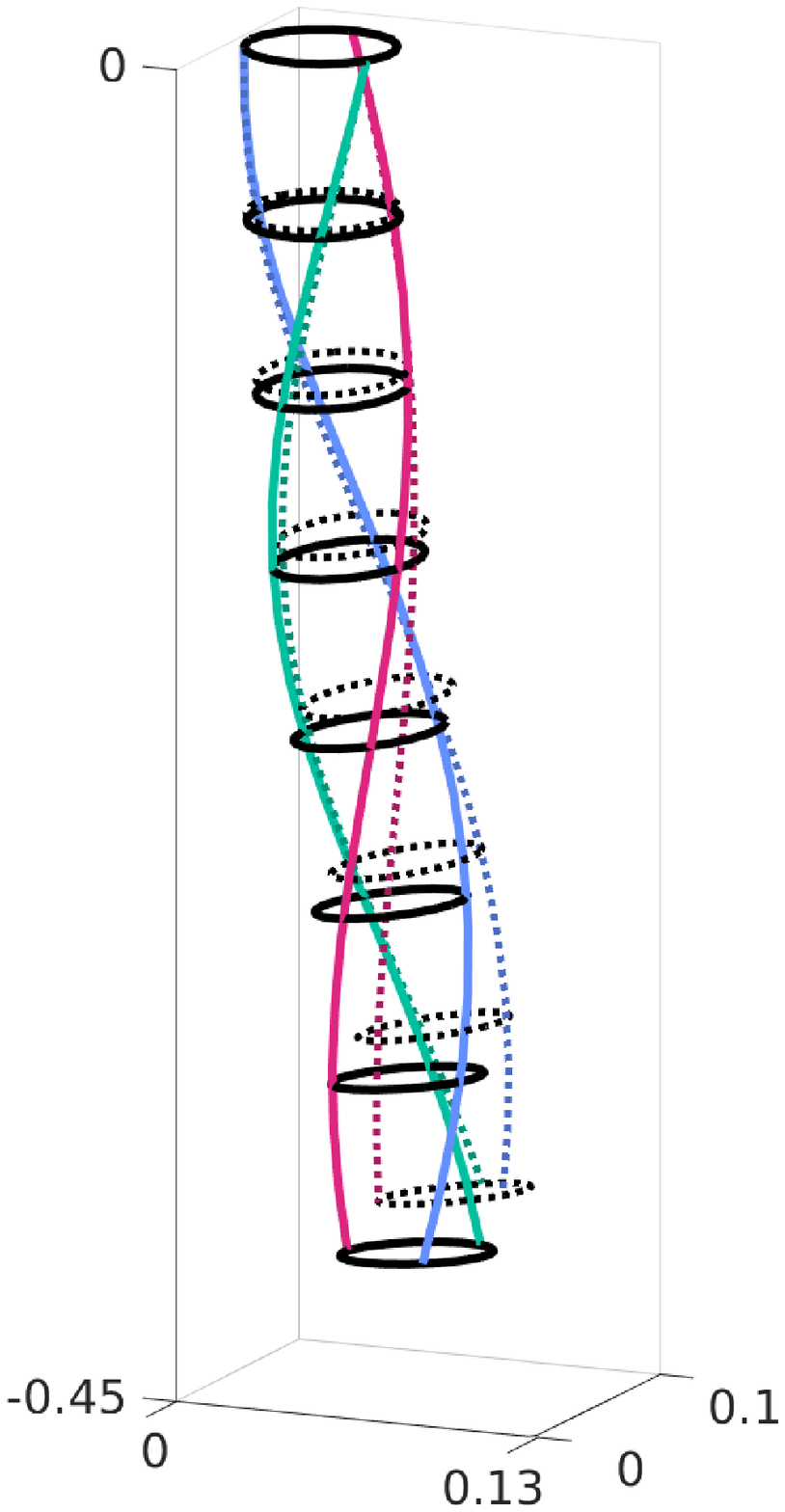}
        \centering
    \end{subfigure}
    \begin{subfigure}[b]{0.32\columnwidth}
        \includegraphics[width=\textwidth]{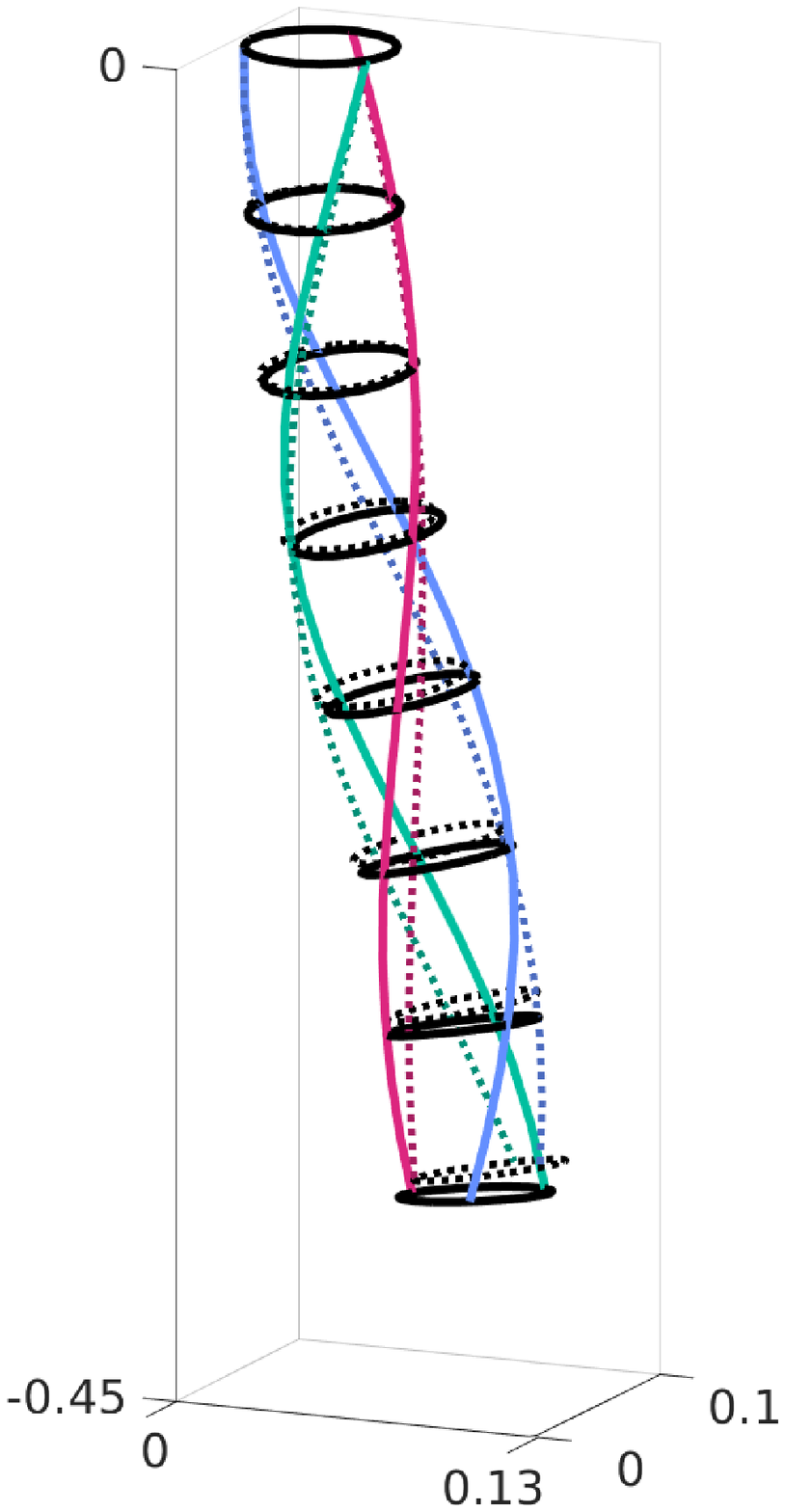}
        \centering
    \end{subfigure}

    \caption{\textbf{Expected vs. measured geometry of a helical arm.} Left to right: red muscle inflated at 103kpa, 241kpa, and 345kpa.}
    \label{fig:helical_scenes}
\end{figure}

\begin{figure*}
    \begin{subfigure}[b]{\columnwidth}
        \includegraphics[width=\textwidth]{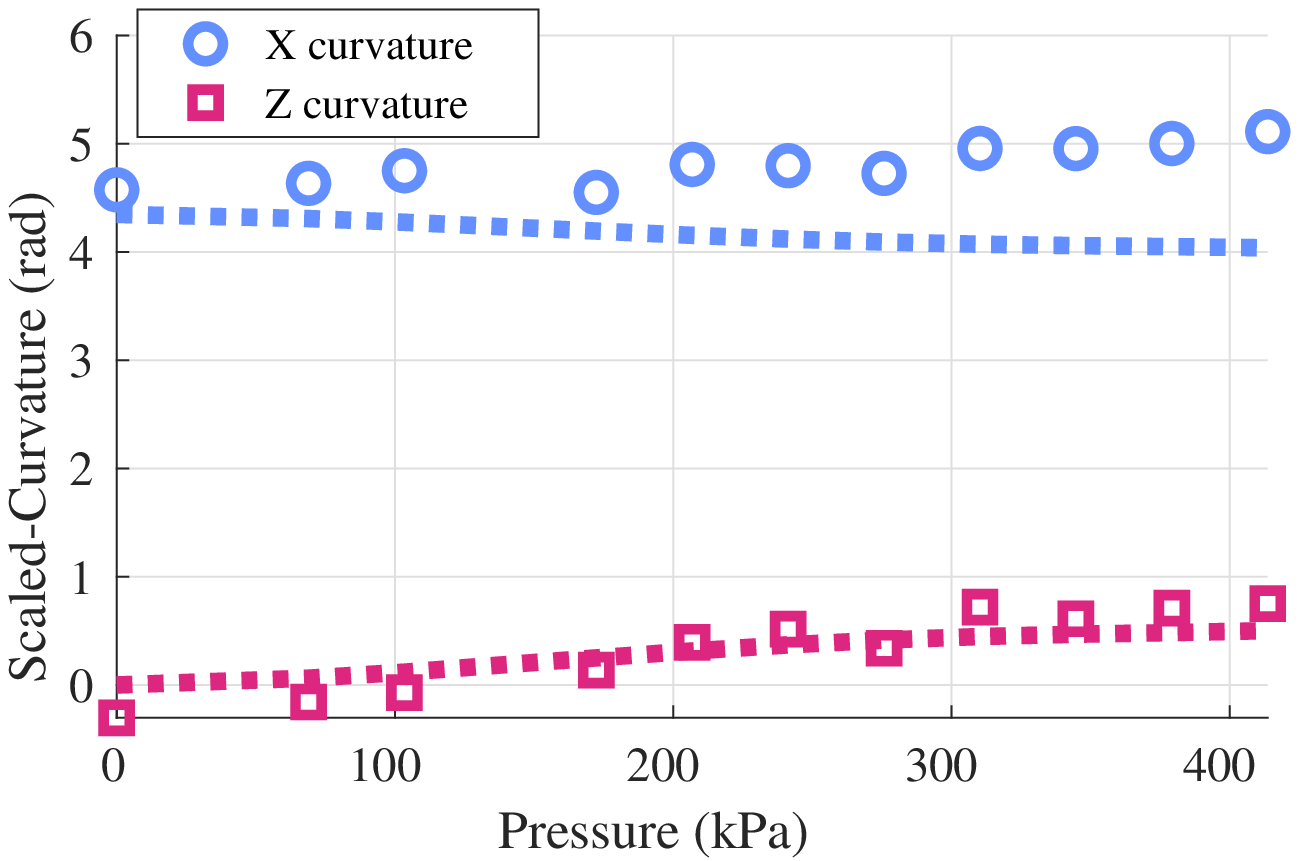}
        \centering
    \end{subfigure}
    \begin{subfigure}[b]{\columnwidth}
        \includegraphics[width=\textwidth]{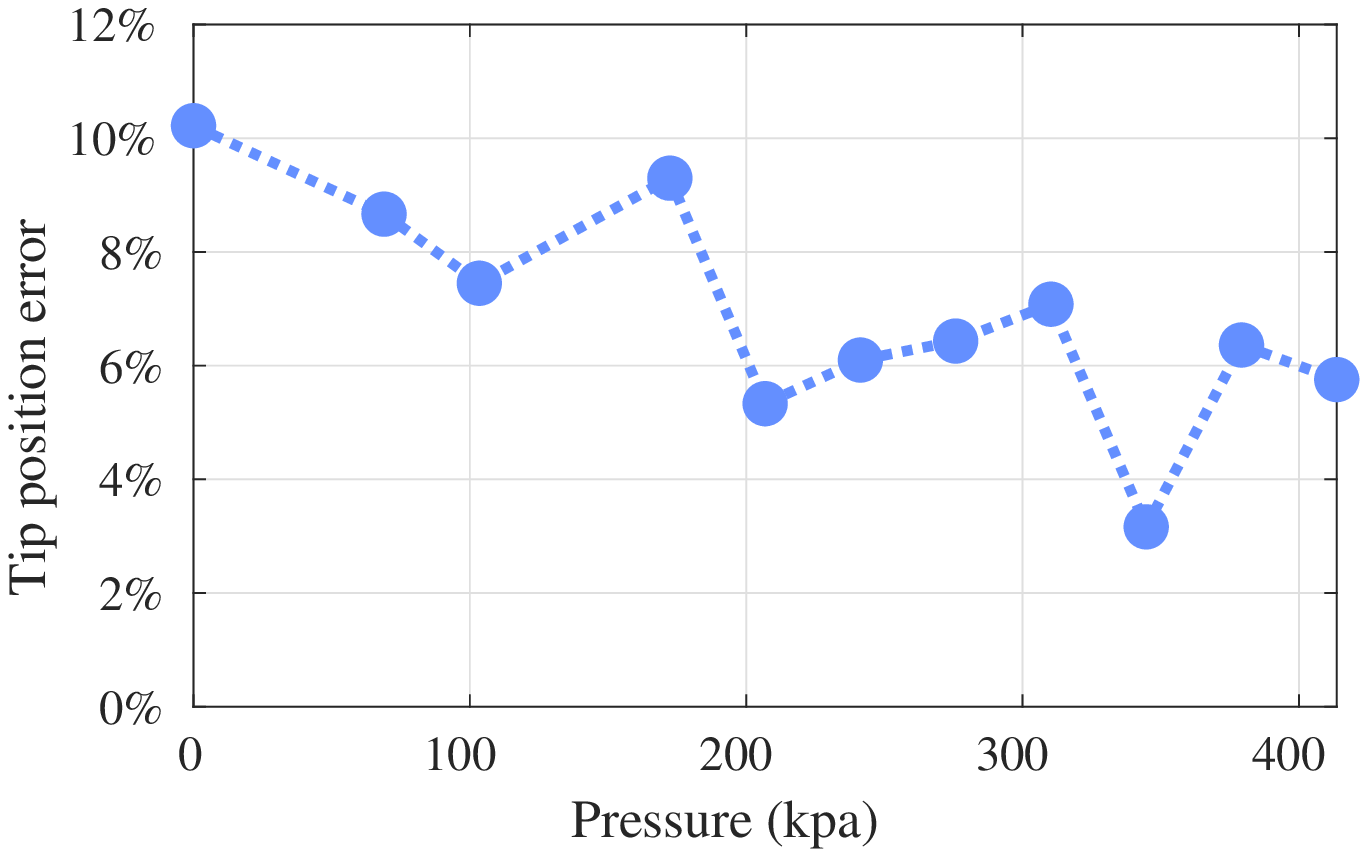}
        \centering
    \end{subfigure}
    \caption{\textbf{Helical arm results model accuracy results.} Left: Model vs. experimentally measured X- and Z-curvature over pressure. Right: Same as before - manipulator tip position error normalized by the length of the manipulator.}
    \label{fig:helical_results}
\end{figure*}

Unlike the planar and spatial arms, helical arm bending is not relegated to a single bending plane. Thus, the effects of gravity could not be prevented, and there are now also two curvature components to consider: the X- and Z-curvature. Figure \ref{fig:helical_results} shows the model accuracy results from actuating a helical arm, where the model has less than 6\% normalized tip position error at maximum inflation..

Prior works on helical manipulators also considered the helix winding radius and helix pitch error. At maximum inflation (380kpa), we measured a winding radius error of 4\%, and a pitch error of 16\%.

\subsection{Sources of Error}
Because our model does not account for external loading, the largest sources of error were external forces such as static friction or gravity, which were most prominent at lower pressures. This can be seen in the planar and helical dataset tip position error trends. Our use of TagSLAM and manually attached Apriltags also likely introduced $\sim$1cm of error to each tag's position estimate. Finally, our assumption of zero torsional stiffness is likely the cause of the incorrectly-trending x-curvature in Fig. \ref{fig:helical_results}. This can be accounted for in future experiments by actuator mounts that prevent them from rotating in-place, as done in \cite{Olson2020AnLoads}.

\section{CONCLUSION}\label{section:conclusion}
In this work, we presented a linear kinematics model that generalizes across all constant curvature and twist manipulators. Using these kinematics, we derived a physics-based mechanics model that is analogous to traditional robot forward kinematics, and generalizes across multiple manipulator designs without the need for re-parameterization. This model was implemented and tested for a wide array of manipulators, and showed less than 10\% of error when using a single characterization across all designs. 

We are excited for future work that applies this model to control and planning of complex constant curvature or twist manipulators that fully exploits the linear forward and inverse kinematics. Potential projects can include finding manipulators and use-cases where the unloaded model is useful for planning and control, as well as extending this model to account for loading while maintaining simplicity. This will bring us another step closer to building continuum manipulators that can reach the full capacity of cephalopod tentacles and elephant trunks.




\section*{ACKNOWLEDGMENT}
This work was supported by the National Science Foundation (NSF) Research Experience for Undergraduates (REU) program. The authors thank Emma Waters for being a wonderful project partner and friend, Luke Raus for his feedback on figure design, and all friends for their support.

\printbibliography


\end{document}